\documentclass[9 pt,conference]{IEEEtran}
\IEEEoverridecommandlockouts
\usepackage{cite}
\usepackage{amsmath,amssymb,amsfonts}
\usepackage{algorithmic}
\usepackage{graphicx}
\usepackage{textcomp}
\usepackage{xcolor}
\usepackage{comment}
\usepackage{subcaption}
\usepackage[super]{nth}
\usepackage{enumitem}
\usepackage{balance}

\def\BibTeX{{\rm B\kern-.05em{\sc i\kern-.025em b}\kern-.08em
    T\kern-.1667em\lower.7ex\hbox{E}\kern-.125emX}}
\begin{document}

\title{High-level Modeling of Manufacturing Faults in Deep Neural Network Accelerators}


\author{\IEEEauthorblockN{Shamik Kundu}
\IEEEauthorblockA{\textit{The University of Texas at Dallas}\\
Dallas, Texas \\
shamik.kundu@utdallas.edu}

\and
\IEEEauthorblockN{Ahmet Soyyiğit}
\IEEEauthorblockA{\textit{University of Missouri} \\
Columbia, Missouri\\
as3ff@mail.missouri.edu}
\and

\IEEEauthorblockN{Khaza Anuarul Hoque}
\IEEEauthorblockA{\textit{University of Missouri} \\
Columbia, Missouri\\
hoquek@missouri.edu}

\and
\IEEEauthorblockN{Kanad Basu}
\IEEEauthorblockA{\textit{The University of Texas at Dallas}\\
Dallas, Texas \\
kanad.basu@utdallas.edu}

}

\maketitle

\pagestyle{plain}

\begin{abstract}
The advent of data-driven real-time applications requires the implementation of Deep Neural Networks (DNNs) on Machine Learning accelerators. Google's Tensor Processing Unit (TPU) is one such neural network accelerator that uses systolic array-based matrix multiplication hardware for computation in its crux. Manufacturing faults at any state element of the matrix multiplication unit can cause unexpected errors in these inference networks. In this paper, we propose a formal model of permanent faults and their propagation in a TPU using the Discrete-Time Markov Chain (DTMC) formalism. The proposed model is analyzed using the probabilistic model checking technique to reason about the likelihood of faulty outputs. The obtained quantitative results show that the classification accuracy is sensitive to the type of permanent faults as well as their location, bit position and the number of layers in the neural network. The conclusions from our theoretical model have been validated using experiments on a digit recognition-based DNN.
\\
\end{abstract}

\begin{IEEEkeywords}
Neural Network Accelerator, Tensor Processing Unit (TPU), Probabilistic Model Checking, Stuck-at Faults.
\end{IEEEkeywords}

\section{Introduction}
\label{sec:intro}

The proliferation of low-latency applications necessitate the implementation of Deep Neural Network (DNN) accelerators on Internet-of-things (IoT) edge devices to enhance system performance.
In this direction, researchers have come up with special-purpose accelerators \cite{jouppi2017datacenter, chen2016eyeriss} to expedite the computationally exhaustive Multiplication-Accumulation (MAC) operation of a DNN, which manifests itself as a performance bottleneck. This has been efficiently circumvented by the concept of systolic array, which is an arrangement of data processing units in a two-dimensional homogeneous grid \cite{kung1982systolic}. Google leveraged this concept to develop Tensor Processing Unit (TPU) as their high performing DNN accelerator \cite{jouppi2017datacenter}. TPU uses a 256 $\times$ 256 systolic array-based matrix multiplication unit, furnishing 15–30$\times$ higher performance and 30–80$\times$ higher performance-per-watt over contemporary CPUs and GPUs \cite{jouppi2017datacenter}. In this paper, this TPU architecture from Google is used as a standard baseline design, although our analysis can be extended to any systolic array-based DNN accelerator.

Any Integrated Circuit (IC) is susceptible to manufacturing defects. Faults in a system can be permanent, which result in an unexpected inference outcome. Fault analysis on an accelerator is a fundamental pillar towards calculating error probability. Prior works on faults in systolic array-based hardware designs were discussed in \cite{kung1983fault,kim1989design,zhang2018analyzing}, where the authors injected faulty MAC units in the systolic array and demonstrated intense performance penalty of the network under faults. Fault mitigation approaches were also explored, either by compromising classification accuracy or by increasing area overhead. 
However, none of these articles focus on a formal analysis to estimate the terminating error probability.

This article focuses on developing a formal model using the Discrete-Time Markov Chain (DTMC) formalism that captures the behavior of a permanent fault in a TPU. The proposed model is analyzed using the probabilistic model checking (PMC)~\cite{kwiatkowska2002prism} technique to reason about the probability of misclassified outcomes. PMC is a formal verification technique that furnishes numerically exact answers to the temporal logic queries by virtue of its exhaustive analysis, in contrast to discrete-event simulations. The quantitative analysis using our model shows that the type of permanent faults (stuck-at-1 vs. stuck-at-0), the position of the permanent faults, and the number of layers in the neural network directly impact the classification accuracy of the DNN accelerator. Such analysis can aid in approximating the sensitivity of the hardware under various fault parameters, and thus can help the designers to adopt the proper fault mitigation techniques, thereby increasing the yield. To the best of our knowledge, this is the \emph{first work} that proposes a formal modeling and quantitative analysis of permanent faults in DNN accelerators.

The rest of the paper is organized as follows. In Section II, the background information on a systolic array-based TPU is presented. Section III presents the modeling of stuck-at faults in the MAC array, and their propagation through the accelerator to the output. Section IV explains the experimental results and validation of our proposed model. Finally, Section V concludes the paper.

\section{Background}
\vspace{-1mm}

A neural network consists of multiple computation layers that extract relevant features from raw input data to perform prediction via the complex analytical architecture. Each layer comprises multiple neurons that perform matrix multiplication, followed by a non-linear activation function. The complex computation of DNNs is efficiently implemented using a systolic array architecture, which is a $N \times N$ homogeneous grid of densely connected MAC units in commercial accelerators like TPU \cite{jouppi2017datacenter}. $MAC_{r,c}$ represents a MAC unit at the intersection of $r^{th}$ row and $c^{th}$ column. Layers are mapped onto the systolic array grid, where each column represents a neuron in a distinct layer. At first, the weight matrix is loaded into the systolic array through the weight memory and the activation inputs are loaded into the activation memory. In the first clock cycle, the unit $MAC_{1,1}$ computes the product between weight element and the corresponding activation input, after which the result is transferred vertically to $MAC_{2,1}$ and the activation input is transferred horizontally to the $MAC_{1,2}$. In the second clock stage, the $MAC_{2,1}$ and the $MAC_{1,2}$ computes the corresponding product. This process is replicated until the accumulators finally add up the products from each $MAC$ column, following which the results move into the activation block to be fed into the subsequent quantization phase to meet specific hardware constraints. The quantized output is eventually provided as activation to the subsequent layer in the network.


\section{Modeling}
\label{sec:modeling}
Let FTPU be a TPU whose systolic array has $R$ rows and $C$ columns while one of its MAC units has a permanent fault. Let this faulty MAC unit be denoted as $FMAC_{r,c}$, which is positioned at row $r$ and column $c$ of the systolic array. Also, let NFTPU be another TPU which has the same attributes as FTPU, but without a permanent fault. We consider a Neural Network (NN) whose attributes and related functions are given below.
\begin{itemize}[leftmargin=*]
\item The NN has $L$ layers, with $N$ neurons in each layer ($1 \leq N \leq R$), where $l$ represents any layer ranging from 1 to $L$.
\item All layers in the network are fully connected and use the same set of weights from the matrix $W_{N \times N}$, where  $w_{i,j}$ represents the weight at the intersection of $i^{th}$ row and $j^{th}$ column.

\item The activations provided to layer $l$, $X(l)$, is defined as:
\begin{equation}
X(l)=\begin{bmatrix}
    x_{1}(l) & x_{2}(l) & \dots   & x_N(l)
\end{bmatrix}^T
\end{equation} 
where any $x_i(l)$ is the activation of neuron $i$ at layer $l$.
\item Let $a$ and $m$ be the row and column indices of $W_{N \times N}$. Also, let the multiplication result of the MAC unit loaded with $w_{a,m}$ be $y_{a,m}(l)$, and let the accumulation result of the same MAC unit be $Y_{a,m}(l)$ during the execution of layer $l$. The $m^{th}$ activation of the next layer, $x_m(l+1)$, is calculated as follows:
\begin{equation}
y_{a,m}(l)=x_a(l)\times w_{a,m}
\end{equation}
\begin{equation}
Y_{a,m}(l)=y_{1,m}+...+y_{a-1,m} + y_{a,m}
\end{equation}
\begin{equation}
x_m(l+1)=\phi(Y_{N,m})
\end{equation}
where $\phi$ is the activation function and $1 \leq a \leq m \leq N$. We considered Rectified Linear Unit (\textit{ReLU}) as the activation function.
\end{itemize}

When a layer with $N$ neurons is being processed, only the MAC units inside the bottom leftmost $N \times N$ square area receive and process activations. We call this zone of MAC units as the effective area. 
Any $MAC_{r,c}$ that satisfies the conditions $(r_{inv} \leq N)$ and $(c\leq N)$ will be inside the effective area, where $r_{inv}=R-r+1$ is defined as the inverse row number. As the effective area changes according to $N$, the activation which arrives to a particular MAC unit also changes. 
To interpret which activation will be processed by a $MAC_{r,c}$, we define a relative row number $rr=r-R+N$. A MAC unit, $MAC_{rr,c}$, calculates $y_{rr,c}$ by multiplying $x_{rr}$ with $w_{rr,c}$, and accumulates $y_{rr,c}$ with $Y_{rr-1,c}$ to produce $Y_{rr,c}$.

We focus on the problem of determining the probability of having error-free output from FTPU if NN is feed-forwarded through it. The proposed model is based on calculating and comparing the results of FTPU and NFTPU for all possible set of activations. The model exhaustively calculates the probabilities depending on the given probability distribution of each activation value inside $X(1)$. We investigate this problem for three different hardware fault cases:\\
\textit{Case 1}: One bit is stuck-at-0 or 1 in the weight register of $FMAC_{r,c}$.\\
\textit{Case 2}: One bit is stuck-at-0 or 1 in the accumulator of $FMAC_{r,c}$.\\
\textit{Case 3}: One bit is stuck-at-0 or 1 in the multiplier of $FMAC_{r,c}$.\\
For all these three cases, we represent the index of the stuck-at bit with $SP$, and type of the stuck-at fault with $ST$ which can be either 0 or 1. Also, we use $SM=2^{SP}$ to represent the stuck-at mask. 

The effect of $FMAC_{r,c}$ should be expressed mathematically for each fault case in order to develop a solution to the problem. This effect is applied on all layers of $NN$ individually as the systolic array is used repetitively for each layer. 
Let us consider Case 1. It is important to note that, when $FMAC_{r,c}$ is not inside the effective area, or when the stuck-at error does not change the weight loaded to it (which means masked), the stuck-at fault won't have any effect. Even if the fault changes the weight, the produced error might be masked after the activation function, or quantization. 



The following equation expresses the fault effect for Case 1, $F_w(l)$, which is an integer added to $Y_{N,c}(l)$ 
after processing layer $l$.

\begin{equation}
F_w(l) = \begin{cases}
{SM} \times x_{rr}(l), &\text{if $C_0 \land\neg C_1$}\\
-({SM}\times x_{rr}(l)), &\text{if $C_0 \land\neg C_2$}\\
0, &\text{otherwise}
\end{cases}
\end{equation}
The conditions in the expression represent:

\begin{itemize}
\item $C_0$ (Effective area): $(r_{inv} \leq N) \land (c\leq N)$
\item $C_1$ (Stuck-at-1 masked): $[(w_{rr,c}\mathbin{\&}SM)\neq 0]\land (ST=1)$
\item $C_2$ (Stuck-at-0 masked): $[(w_{rr,c}\mathbin{\&}SM)=0] \land (ST=0)$
\end{itemize}

where $\land$, $\neg$, and $\mathbin{\&}$ denotes the logical-and, logical-negation, and bit-wise-and operators, respectively. The condition $C_0$ states that $FMAC_{r,c}$ is going to be inside the effective area. Depending on the value of $w_{rr,c}$, the fault can be masked inside $FMAC_{r,c}$, which is stated as condition $C_1$ for stuck-at-1, and $C_2$ for stuck-at-0. 

Case 2 has two different effects. The first one is simply altering the value $Y_{rr,c}(l)$. 
The second effect can happen only if the accumulator of $FMAC_{r,c}$ has stuck-at-1 fault. This effect is caused by triggering all MAC units with the same clock, which forces them to produce an output value at every cycle. $FMAC_{r,c}$ produces a value equal to $SM$, unlike non-faulty MAC units which produce a zero when they are not processing activations. The produced $SM$ values go down through the column $c$ and the ones that reach to the bottom accumulator are added to $Y_{N,c}(l)$. We define this as leaking effect.

The error representing Case 2, $F_a(l)$ can be expressed as follows:

\begin{equation}
F_a(l) = \begin{cases}
(2N-r_{inv}+1)\times SM, &\text{if $C_3 \lor (C_0 \land \neg C_4)$}\\
(2N-r_{inv})\times SM, &\text{if $C_0 \land C_4$}\\
-SM, &\text{if $C_0 \land \neg C_5$}\\
0, &\text{otherwise}
\end{cases}
\end{equation}

Apart from $C_0$, other conditions in the equation indicate:
\begin{itemize}
\item $C_3$ (Upper effective area): \\$(ST=1) \land (N<r_{inv}\leq 2N) \land (c\leq N)$
\item $C_4$ (Stuck-at-1 masked): $(ST=1)\land (Y_{rr,c}(l)\mathbin{\&}SM)\neq 0  $
\item $C_5$ (Stuck-at-0 masked): $(ST=0)\land (Y_{rr,c}(l)\mathbin{\&}SM)= 0$
\end{itemize}
where $\lor$ represents logical-or operator. The condition $C_3$ represents an area of systolic array where $FMAC_{r,c}$ does not process activation, but it effects the $Y_{N,c}(l)$ due to the leaking effect. Please note that, for $N$ neurons, TPU requires $2N+1$ cycles to finish the systolic array operation. $C_4$ and $C_5$ represent the situation where stuck-at fault does not change $Y_{rr,c}(l)$. 

Lastly, we will discuss Case 3. The leaking effect behaves in the same way as in Case 2, since the output of multiplier is connected to the accumulator inside the MAC unit. 
The equation of the total error being produced for Case 3 defined as $F_m(l)$ is as follows:
\begin{equation}
F_m(l) = \begin{cases}
(2N-r_{inv}+1)\times SM, &\text{if $C_3 \lor (C_0 \land \neg C_6)$}\\
(2N-r_{inv})\times SM, &\text{if $C_0 \land C_6$}\\
-SM, &\text{if $C_0 \land \neg C_7$}\\
0, &\text{otherwise}
\end{cases}
\end{equation}
The conditions $C_6$ and $C_7$ used in the expression denote:
\begin{itemize}
\item $C_6$ (Stuck-at-1 masked): $(ST=1)\land (y_{rr,c}(l)\mathbin{\&}SM)\neq 0  $
\item $C_7$ (Stuck-at-0 masked): $(ST=0)\land (y_{rr,c}(l)\mathbin{\&}SM)= 0$
\end{itemize}

Using the equations $(5)$, $(6)$, and $(7)$, we model the FTPU and NFTPU in the presence of faults. The model has two main parts \textemdash input selection and result calculation. 
The input selection is modeled using a probabilistic automaton, specifically using DTMC formalism. The result calculation is modeled as a special case of probabilistic automaton to demonstrate respective deterministic behavior.

\noindent \textbf{Definition 1 (IS-DTMC).}
\textit{The process of selecting an input value for a single neuron of the first layer can be defined as a DTMC $D_x=(S,s_{init},TL,P,L)$, where:
\begin{itemize}[leftmargin=*]
\item[--] $S=\{s_0,s_1,...,s_v\}$, where each $s_i\in S$ represents an input $i>0$;
\item[--] $s_{init}=\{s_0\}$, is the initial state;
\item[--] $TL=\{i_1,i_2,...,i_v\}$ is a set of transition labels.
\item[--] $P: S\times S \to [0,1]$ is a transition probability matrix such that $\sum_{s'\in S}p(s,s')=1$, representing probability distribution of inputs.
\item[--] $L:S \to 2^{AP}$ is a function to label each state $s_i \in S$ from the set of Atomic Propositions $AP =\{not\_selected,selected\}$;
\end{itemize}
}

\begin{figure}
\centering
\begin{subfigure}{0.17\textwidth}
  \includegraphics[width=\linewidth]{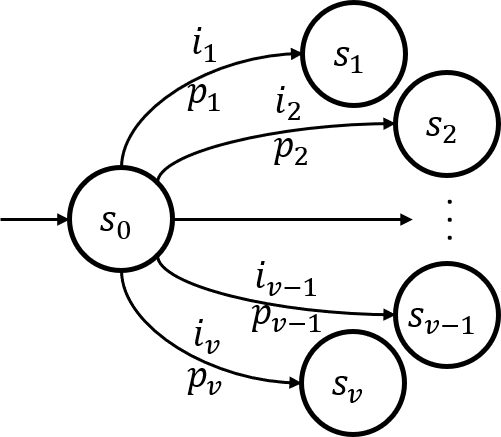}
  \caption{}
\end{subfigure}%
~
\begin{subfigure}{0.23\textwidth}
  \includegraphics[width=\linewidth]{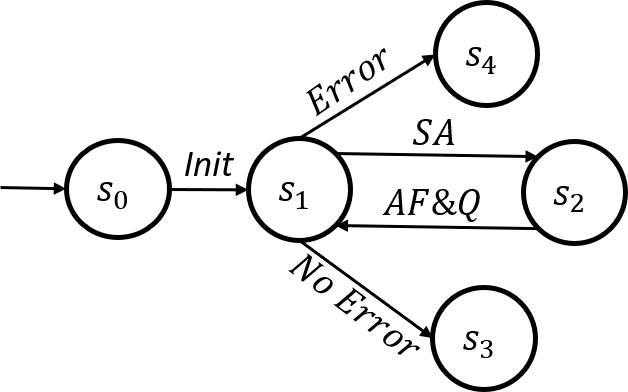}
  \caption{}
\end{subfigure}%

\caption{Representation of (a) IS-DTMC and (b) TPU-FA automaton.}
\label{fig2}
\end{figure}

Visual illustration of IS-DTMC is given in  Figure~\ref{fig2}a, where state $s_0$ denotes that the input is not selected yet, and all other states represent the selected input values. Each transition has probability $p_z$ for selecting input $i_z$, where $1 \leq z \leq v$. An IS-DTMC will have $v=2^b$ transitions if $b$ bits are used to represent the input neuron values in the TPU. 
The transition probabilities in a uniform distribution can be represented as $p_1=p_2=...=p_m=1/v$. A  single neuron can be modeled using a IS-DTMC model. To model $N$ number of neurons, $N$ number of IS-DTMC models are required which are synchronized using the transition label. 
The selected input values are then passed to the output calculation part represented using the TPU-FA (TPU-Fault Analysis) automaton.


\noindent \textbf{Definition 2 (TPU-FA Automaton).} \textit{The execution steps to analyze the outputs of NFTPU and FTPU can be defined as a special case of the probabilistic automaton $E=(S,s_{init},TL,P,L)$ where:
\begin{itemize}[leftmargin=*]
\item[--] $S=\{s_0,s_1,s_2,s_3,s_4\}$, represents the state of execution;
\item[--] $s_{init}=\{s_0\}$, is the initial state; 
\item[--] TL = \{Init,SA,AF\&Q,No\ Error,Error\} is a set of transition labels;
\item[--]$P: S\times S \to [0,1]$ is a transition probability matrix, where probabilities are set to 1, since the execution steps are deterministic.
\item[--] $L:S \to 2^{AP}$ is a function to label each state $s_i \in S$ from the set of Atomic Propositions $AP =\{begin, ready, calculating, end\}$;
\end{itemize}}

The TPU-FA automaton $E$ represents the execution cycle of a TPU as illustrated in Figure~\ref{fig2}b. Firstly, it takes the transition $Init$ to initialize inputs chosen by the IS-DTMCs. At state $s_1$, if the feedforward has not completed yet, it takes the transition \textit{SA (Systolic Array)} to state $s_2$ and processes the next layer by doing the systolic array calculations for both the NFTPU and the FTPU. For calculating the result of FTPU, it uses the equations (5), (6) or (7), depending on the chosen fault case. Afterwards, it goes back to $s_1$ by transition $AF\&Q(Activation\ Function\ \&\ Quantization)$ while forwarding the results through the activation function followed by the quantization. When all the layers are processed, the TPU-FA automaton compares the results of both TPUs in state $s_1$. If the results are equal, then the automaton takes the transition $No\ Error$ leading to state $s_3$, if not, then it takes the transition $Error$ leading to $s_4$. 

For modeling and automated analysis of the proposed model, we use the probabilistic model checking (PMC) technique~\cite{kwiatkowska2007stochastic}, more specifically, the PRISM model checker~\cite{kwiatkowska2002prism}. 
We are interested in analyzing the case where the outputs from NFTPU and FTPU are not equal. This property can be formalized in Probabilistic Computation Tree Logic (PCTL)~\cite{hansson1994logic} as  $P_{=?} [\ F\ ``error"\ ]$, where ``error" represents state $s_4$ and $F$ represent the \emph{eventually} operator. 








\section{Experimental Results}
\subsection{Experimental Setup}
\label{setup}
In this section, we present the modeling results and validate them. 
For the experiments, we consider a systolic array prototype with a grid of $4 \times 4$ MAC units. Each of the weight and the activation registers have a length of 4 bits, thereby furnishing a multiplier output of length 8 bits and an accumulator output of length 10 bits. 
For the DTMC model, the weights are chosen randomly for each neuron. 
Given a fixed width of weight and activation registers, the total number of states and transitions in the model $\mathcal{M}$ (after the parallel composition) depends on the number of layers. 

\subsection{Modeling Results}
\label{results}

\subsubsection{Impact of Number of Layers on Error Probability} 
\begin{figure*}
\centering
\begin{subfigure}{0.33\textwidth}
  \includegraphics[width=\linewidth]{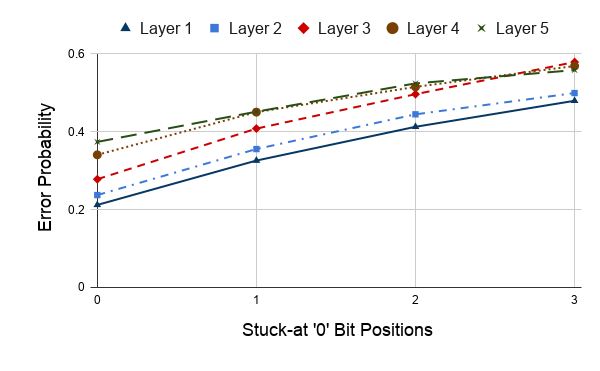}
  \caption{Sensitivity of stuck-at-0 bit positions.}
  \label{fig:sa0wdup}
\end{subfigure}%
~
\begin{subfigure}{0.33\textwidth}
  \includegraphics[width=\linewidth]{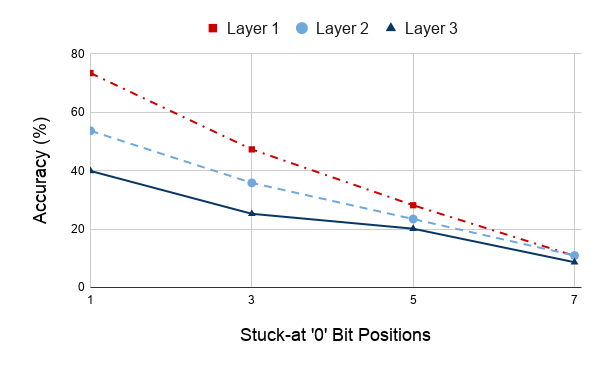}
  \caption{Validation: Accuracy vs. bit positions (s-a-0).}
  \label{fig:versa0}
\end{subfigure}%
~
\begin{subfigure}{0.33\textwidth}
  \includegraphics[width=\linewidth]{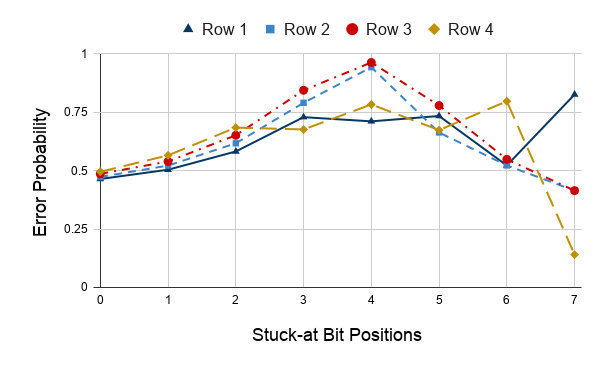}
  \caption{Sensitivity of layers for fault in multiplier.}
  \label{fig:sa1mul}
\end{subfigure}%

\begin{subfigure}{0.33\textwidth}
  \includegraphics[width=\linewidth]{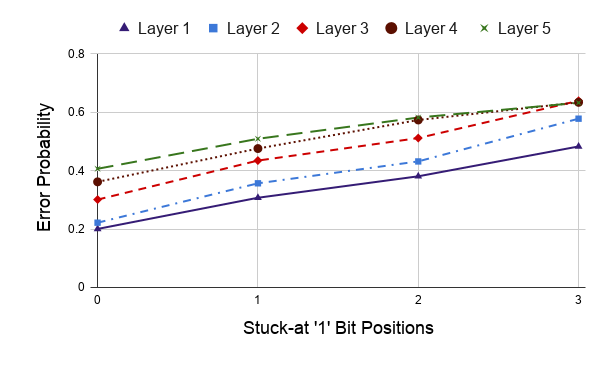}
  \caption{Sensitivity of stuck-at-1 bit positions.}
  \label{fig:sa1wdup}
\end{subfigure}%
~
\begin{subfigure}{0.33\textwidth}
  \includegraphics[width=\linewidth]{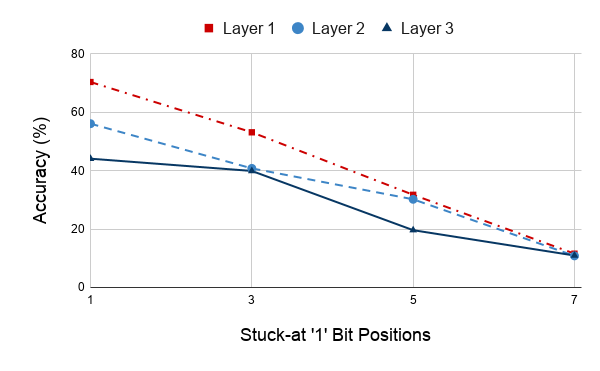}
  \caption{Validation: Accuracy vs. bit positions (s-a-1).}
  \label{fig:versa1}
\end{subfigure}%
~
\begin{subfigure}{0.33\textwidth}
  \includegraphics[width=\linewidth]{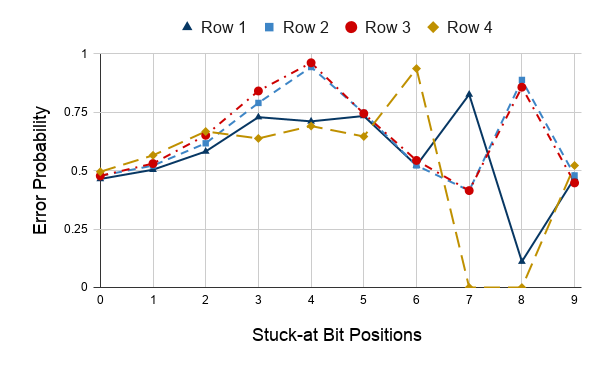}
  \caption{Sensitivity of layers for fault in accumulator.}
  \label{fig:sa1add}
\end{subfigure}%

\caption{Analyzing output behavior for diverse fault parameters.}
\label{fig:total}
\vspace{-4mm}
\end{figure*}

In this experiment, the probability of having an error in the output is observed for a model with incremental layers in the network, by inducing faults at individual bit positions of the weight register. Figure~\ref{fig:sa0wdup} and~\ref{fig:sa1wdup} exhibit consistent increase in error probability with the increase in number of layers from 1 to 5, as well as with the variation of stuck-at positions in the weight from the Least Significant Bit (LSB) `0' through the Most Significant Bit (MSB) `3' for a distinct layer in the DNN model, for both stuck-at-0 and stuck-at-1 respectively. As the significance of the bit position increases from the LSB to the MSB, the deviation from the accurate value consistently enhances, thereby increasing the probability of an erroneous output. Furthermore, since all the layers in the model are fully connected, a single unmasked fault in a MAC unit of a particular layer affects all the connected nodes in the subsequent layers, thus enhancing output error probability. 

\subsubsection{Impact of Faulty MAC Position on Error Probability} 

In this experiment, stuck-at-1 faults are introduced in each bit of the multiplier and accumulator for individual MAC units in rows one through four, all positioned in the first column of the systolic array. 
Figure~\ref{fig:sa1mul} and Figure~\ref{fig:sa1add} represent the variation of error probability for individual faulty MAC row locations, considering faults in the multiplier and the accumulator respectively. As the row position of a faulty MAC unit varies from one through four, it approaches nearer to final accumulator output. In this scenario, the generated error from the fault becomes difficult to mask due to reduced effect of ensuing MAC units, thereby gradually increasing the output error probability. However, after a certain bit position in both the graphs, the variation does not follow the pattern. This can be primarily subjected to overflow issues, which arise as a result of leaking effect from the stuck-at-1 fault in higher order bits. For example, let us consider \textit{Case 2} where $N=4$, $FMAC_{1,1}$, $SM=2^8$, and the condition $(C_0 \land C_4)$ is satisfied. In this case, $F_m(l)$ will be equal to $(2N-r_{inv})*SM=(8-4)*2^8=2^{10}$, which will overflow and become 0 as the accumulator can only accommodate ten bits. This effect can be seen in Figure~\ref{fig:sa1add}, for a specific set of random weights, where the error probability reduces to 0.125 for Row 1 when $SM=2^8$. However, bit truncation operation in the quantization phase also plays a trivial role in this phenomenon. In Figure~\ref{fig:sa1add}, although overflow does not occur for $SM=2^7$, the fault gets masked as a result of quantization and the error probability plummets close to zero. On the other hand, for $SM=2^9$, even though there is an overflow effect, the error probability increases, because the cumulative effect of the fault throughout the register cannot entirely be masked by bit truncation.\\

\vspace{-4mm}

\subsection{Validation Results}

The observations from the proposed hypothesis are validated by simulating a neural network using \textit{Keras} in \textit{Python}. The model is trained on MNIST handwritten digit database of 60,000 images. 
A low classification accuracy of a model is illustrative of a high output error probability and vice-versa. Both stuck-at-0 and 1 faults are induced in the weights and the mean accuracy is plotted, as shown in Figure~\ref{fig:versa0} and Figure~\ref{fig:versa1} respectively. 

From the plots, it is observed that as the faulty bit position varies from LSB `1' to MSB `7', the impact on prediction accuracy increases, thereby corroborating our observation on the impact of faulty bit position. The increase in number of faulty layers in the network also reduces the accuracy significantly, as observed from the hypothesised fault model. The observation on faulty MAC position cannot be directly verified by the validation model as it cannot access the accumulated output of each MAC unit in the systolic array individually.
However, a fault in the multiplier can be translated to a fault in the weight matrix, the impact of which have been validated as shown in Figure ~\ref{fig:versa0} and Figure ~\ref{fig:versa1}. The classification model has an initial accuracy of $97.72\%$ without any stuck-at faults. With the induction of fault in the seventh bit position, the accuracy reduces to an average of $10.15\%$ and $11.13\%$ for stuck-at-0 and stuck-at-1 respectively. Hence, the obtained results are coherent with the analytics derived from the formal model, thereby justifying the correctness of the model.


\section{Conclusion}
The proposed formal model and an extensive analysis of permanent faults in a TPU provide an insight into its expected behavior. The obtained results show that the number of layers in a neural network, the type of faults and their location have a direct impact on the probability of misclassification. The proposed high-level model can be used by the hardware designers in adopting proper fault mitigation techniques for improving the yield. Note, in addition to the number of layers, the size of the model (the number of states) also depends on the size of activation and weight registers. 
Since PRISM includes multiple model checking engines enabling the probabilistic verification of models of up to $10^{10}$ states, it suffers from state space explosion problem. However, different state abstraction techniques exist to overcome this issue. In the future, we plan to model the effect of temporal faults in TPUs (such as soft-errors~\cite{hoque2013early}).

\balance


\begin{thebibliography}{10}
\providecommand{\url}[1]{#1}
\csname url@samestyle\endcsname
\providecommand{\newblock}{\relax}
\providecommand{\bibinfo}[2]{#2}
\providecommand{\BIBentrySTDinterwordspacing}{\spaceskip=0pt\relax}
\providecommand{\BIBentryALTinterwordstretchfactor}{4}
\providecommand{\BIBentryALTinterwordspacing}{\spaceskip=\fontdimen2\font plus
\BIBentryALTinterwordstretchfactor\fontdimen3\font minus
  \fontdimen4\font\relax}
\providecommand{\BIBforeignlanguage}[2]{{%
\expandafter\ifx\csname l@#1\endcsname\relax
\typeout{** WARNING: IEEEtran.bst: No hyphenation pattern has been}%
\typeout{** loaded for the language `#1'. Using the pattern for}%
\typeout{** the default language instead.}%
\else
\language=\csname l@#1\endcsname
\fi
#2}}
\providecommand{\BIBdecl}{\relax}
\BIBdecl

\bibitem{jouppi2017datacenter}
Jouppi \emph{et~al.}, ``In-datacenter performance analysis of a tensor
  processing unit,'' in \emph{2017 ACM/IEEE 44th Annual International Symposium
  on Computer Architecture (ISCA)}, pp. 1--12.

\bibitem{chen2016eyeriss}
Chen \emph{et~al.}, ``Eyeriss: An energy-efficient reconfigurable accelerator
  for deep convolutional neural networks,'' \emph{IEEE Journal of Solid-State
  Circuits}, vol.~52, no.~1, pp. 127--138, 2016.

\bibitem{kung1982systolic}
H.-T. Kung, ``Why systolic architectures?'' \emph{IEEE computer}, vol.~15,
  no.~1, pp. 37--46, 1982.

\bibitem{kung1983fault}
Kung \emph{et~al.}, ``Fault-tolerance and two-level pipelining in vlsi systolic
  arrays,'' Carnegie-Mellon Univ Pittsburgh PA Dept of Computer Science, Tech.
  Rep., 1983.

\bibitem{kim1989design}
Kim \emph{et~al.}, ``On the design of fault-tolerant two-dimensional systolic
  arrays for yield enhancement,'' \emph{IEEE Transactions on Computers},
  vol.~38, no.~4, pp. 515--525, 1989.

\bibitem{zhang2018analyzing}
Zhang \emph{et~al.}, ``Analyzing and mitigating the impact of permanent faults
  on a systolic array based neural network accelerator,'' in \emph{2018 IEEE
  36th VLSI Test Symposium (VTS)}, pp. 1--6.

\bibitem{kwiatkowska2002prism}
Kwiatkowska \emph{et~al.}, ``{PRISM: Probabilistic symbolic model checker},''
  in \emph{International Conference on Modelling Techniques and Tools for
  Computer Performance Evaluation}.\hskip 1em plus 0.5em minus 0.4em\relax
  Springer, 2002, pp. 200--204.

\bibitem{kwiatkowska2007stochastic}
------, ``Stochastic model checking,'' in \emph{International School on Formal
  Methods for the Design of Computer, Communication and Software
  Systems}.\hskip 1em plus 0.5em minus 0.4em\relax Springer, 2007, pp.
  220--270.

\bibitem{hansson1994logic}
Hansson \emph{et~al.}, ``A logic for reasoning about time and reliability,''
  \emph{Formal aspects of computing}, vol.~6, no.~5, pp. 512--535, 1994.

\bibitem{hoque2013early}
Hoque \emph{et~al.}, ``Early analysis of soft error effects for aerospace
  applications using probabilistic model checking,'' in \emph{International
  Workshop on Formal Techniques for Safety-Critical Systems}.\hskip 1em plus
  0.5em minus 0.4em\relax Springer, 2013, pp. 54--70.

\end{thebibliography}


\bibliographystyle{IEEEtran}
\end{document}